\documentclass{IOS-Book-Article}
\usepackage{graphicx}
\usepackage{mathptmx}
\usepackage{soul}\setuldepth{article}
\def\hb{\hbox to 11.5 cm{}}

\begin{document}

\pagestyle{headings}
\def\thepage{}
\begin{frontmatter}              

\title{A Hazard-Informed Data Pipeline for Robotics Physical Safety}

\markboth{}{March 2026\hb}

\author{\fnms{Alexei} \snm{Odinokov}},
\author{\fnms{Rostislav} \snm{Yavorskiy}%
\thanks{Corresponding Author: Rostislav Yavorskiy, RYavorsky@gmail.com.}}

\runningauthor{A. Odinokov, R. Yavorskiy}
\address{SafePi.ai, Madrid, Spain}

\begin{abstract}
This report presents a structured Robotics Physical Safety Framework based on explicit asset declaration, systematic vulnerability enumeration, and hazard-driven synthetic data generation. The approach bridges classical risk engineering with modern machine learning pipelines, enabling safety envelope learning grounded in a formalized hazard ontology.
The key contribution of this framework is the alignment between classical safety engineering, digital twin simulation, synthetic data generation, and machine learning model training.
\end{abstract}

\end{frontmatter}

\section{Introduction}

Robotic systems increasingly operate in close proximity to humans, critical infrastructure, and natural ecosystems.
Traditional robotics safety focuses primarily on deterministic failure modes. However, modern Physical AI systems exhibit complex, adaptive behavior, where risks can emerge from large-scale interactions rather than isolated faults. This demands a more systematic engineering approach.

As Physical AI systems scale across industries, robotics safety must integrate formal hazard reasoning with modern data-driven techniques. The proposed Hazard-Informed Synthetic Data Pipeline provides a reproducible methodology for embedding safety into both simulation and ML training workflows.

Rather than training models to recognize accidents after they occur, we propose training them within a formally declared universe of potential harm.

Safety, in this view, is not only about preventing failure, it is about systematically modeling what must be protected, how it can be exposed, and how harm may emerge.

\subsection{Deterministic vs. Emergent Harm}

Robotic systems present a complex and evolving landscape of safety challenges. To effectively model and mitigate these risks, it is useful to categorize the potential for harm into two fundamental types: deterministic harm and emergent harm.

Deterministic harm is characterized by a clear, traceable cause-and-effect chain. These failures are typically reproducible and can often be anticipated through rigorous pre-deployment testing. Consequently, they are amenable to established mitigation techniques drawn from classical engineering and software safety. These methods include high-fidelity simulation, hardware redundancy, formal verification of control software, and system certification against defined standards. Common examples in this category include:

\begin{itemize}
\item A robotic arm exceeding joint limits and crushing an object

\item Sensor failure leading to incorrect control input and collision

\item Brake actuator overheating preventing timely stop

\item Software defect triggering unintended motion
\end{itemize}

In contrast, emergent harm arises not from the failure of a single component, but from the complex, nonlinear interactions within a system operating at scale or within an uncontrolled environment. Here, the path to harm is diffuse and system-level. Each individual robot and subsystem may be functioning precisely as designed, yet the collective behavior produces an unsafe or undesirable outcome. This class of harm is far more difficult to predict and model using traditional techniques. Examples include:

\begin{itemize}
\item Individually correct warehouse robots creating collective deadlocks

\item Delivery robots altering pedestrian flow in urban areas

\item Agricultural robots over-optimizing yield while degrading soil biodiversity

\end{itemize}

This distinction motivates a comprehensive safety framework capable of systematically modeling both localized, component-level risks and the diffuse, systemic risks that emerge from collective behavior. Addressing emergent harm requires a shift from purely first-principles models toward data-driven approaches that can capture the unpredictable dynamics of real-world, multi-agent systems.

\subsection{The Role of Synthetic Data in Mitigating Emergent Harm}

While foundational machine learning models offer powerful generalization capabilities, their direct application to safety-critical robotic systems is fraught with challenges. These general models are typically trained on vast but static datasets that may not capture the unique constraints, interaction dynamics, and long-tail events of a specific deployment scenario. Fine-tuning these models to a particular operational context is therefore essential, and synthetic data generation is emerging as a critical tool for this purpose.

The complexity of emergent harm necessitates the exploration of scenarios that are rare or nonexistent in real-world data. Synthetic data provides a controllable and scalable sandbox for this exploration. By simulating multi-agent interactions under a wide array of conditions one can generate rich datasets that specifically target potential failure modes. For instance, one can procedurally generate thousands of warehouse configurations to train a model to recognize precursors to a robot deadlock, or simulate years of pedestrian-robot interaction in a matter of hours to identify potential safety hazards in a new urban delivery system.

This approach enables the creation of ``digital twins'' of the target environment, where models can be iteratively refined and stress-tested against emergent phenomena before any physical deployment. By fine-tuning general models on this high-fidelity synthetic data, we imbue them with an inductive bias toward safety in that specific context. The model learns not just the general rules of navigation, but the subtle dynamics of how its own actions, in concert with others, can lead to undesirable system-level states. In essence, synthetic data bridges the gap between the broad but shallow knowledge of a general model and the deep, context-specific understanding required to ensure the safe and reliable operation of complex, multi-robot systems in the real world.

\subsection{Literature Overview}

The rapid integration of artificial intelligence into physical systems has created an urgent need to reconsider what safety means when AI operates beyond the digital realm. Unlike traditional software systems where failures result in data loss or service interruption, failures in physical AI systems can cause direct bodily harm, property damage, and catastrophic accidents. This review examines several papers published between 2021 and 2026 that collectively trace the evolution of safety thinking in physical AI, revealing a field in transition from narrow, hardware-focused safety concepts toward holistic frameworks that encompass psychological, cyber-physical, and embodied dimensions.

Martinetti et al. \cite{martinetti2021redefining} provide a foundational critique in ``Redefining Safety in Light of Human-Robot Interaction,'' arguing that current standards and regulations operate with an dangerously narrow conception of safety. Their analysis reveals that traditional safety definitions, rooted in directives like the EU Machinery Directive 2006/42/EC, exclusively address physical risks such as mechanical hazards or chemical exposure. However, the integration of artificial intelligence into robotic systems fundamentally challenges this limited view. The authors demonstrate that modern collaborative robots (cobots) engage in multiple interaction modalities simultaneously—physical collaboration in shared workspaces, social interaction that evokes emotional responses, and cognitive engagement that affects worker psychology. Through the compelling case study of pipeline inspection robots, they illustrate how operator cognitive load, interface ergonomics, and the potential for hardware overheating create safety pathways that transcend traditional physical risk assessments. The paper's central contribution is a multidimensional safety framework comprising five interconnected dimensions: physical interaction, social-psychological interaction (including trust, anthropomorphism, and cognitive load), cybersecurity (where compromised nodes can cause physical harm), temporal safety (where harms manifest long after interaction), and societal dimensions (including deskilling and workforce displacement).

Yang et al. \cite{yang2022physical} approach physical safety from a complementary perspective in their survey of IoT equipment security, addressing a critical blind spot in the literature: the physical vandalism and theft of smart devices. While cybersecurity receives extensive attention, the authors note that only approximately four percent of security research considers physical attacks where adversaries simply steal or destroy equipment. Their systematic review of 219 papers reveals that IoT equipment faces distinct physical threats based on deployment context, infrastructure like power lines and smart meters face partial theft and tampering, while mobile devices like smartphones and UAVs face complete theft risks. The paper introduces a useful taxonomy of countermeasures: circuit and system design approaches (such as physical unclonable functions and power frequency signal injection), additional sensing devices (cameras, RFID, vibration sensors), biometric and behavioral analysis (using embedded sensors to detect abnormal usage patterns), and tracking technologies.

Salhab et al. \cite{salhab2024systematic} offer a comprehensive systematic review of AI safety, analyzing 701 studies published between 2018 and 2024 to identify the core concerns shaping contemporary safety research. Their analysis reveals that AI safety clusters around three primary investigation areas: learning technique safety, verification and validation methods, and autonomy management. For supervised learning, safety concerns center on data quality, model complexity, and generalization to out-of-distribution samples, problems that can cause models to fail unpredictably when deployed beyond training conditions. Reinforcement learning, increasingly central to robotics, presents the most complex safety challenges: safe exploration (how agents learn without causing harm), resistance to distributional shifts, avoidance of undesirable side effects, and prevention of reward hacking where agents manipulate their environment to maximize rewards unsafely. The paper systematically catalogs the main safety properties demanding attention: robustness against adversarial attacks and distributional shift, reliability in sustained operation, explainability and interpretability (the ``black box'' problem), fairness and bias mitigation, and transparency.

Jalali Alenjareghi et al. \cite{alenjareghi2024safe} provide a focused examination of how AI can enhance risk assessment methodologies for human-robot collaboration, employing a systematic PRISMA review of 345 papers. Their analysis reveals that traditional risk assessment methods such as Process Failure Mode and Effects Analysis (PFMEA), Hazard and Operability Study (HAZOP), and Fault Tree Analysis (FTA), possess inherent limitations when applied to collaborative robotics. PFMEA systematically identifies process failures but inadequately captures human-robot dynamics; HAZOP provides thorough hazard identification but remains resource-intensive and design-dependent; FTA models complex system failures but often overlooks human factors and relies on historical data that may not predict novel AI behaviors. The paper demonstrates how AI integration can address these limitations through real-time monitoring, predictive analytics, and anomaly detection. The authors propose a hybrid approach combining multiple risk assessment methods with image processing capabilities to enhance risk detection. 

Farahmand and Neu \cite{farahmand2025ai} bridge computer science and engineering perspectives in a technical note that fundamentally reframes how AI safety should be conceptualized for physical infrastructure systems. Their central insight is that computer scientists and engineers approach safety from fundamentally different paradigms. Computer scientists seek formally verifiable, perfectly safe systems through mathematical logic, while engineers adopt probabilistic approaches that accept calculated risks within defined safety margins.  The paper concerns the entanglement of safety and security in connected systems, noting that adversaries no longer need physical access to cause physical harm, they need only compromise a single network node. This insight positions AI safety as inherently requiring both engineering probabilistic thinking and cybersecurity awareness.

Khan et al. \cite{khan2025safety} address the challenge of embedding safety awareness into LLM-driven robot task planning through their SAFER (Safety-Aware Framework for Execution in Robotics) framework. The paper identifies a fundamental limitation of single-LLM planners: context window constraints force trade-offs between maintaining task history for coherence and preserving critical safety constraints, with LLMs naturally prioritizing task completion and efficiency over risk mitigation. Their solution employs multi-LLM collaboration where a dedicated Safety Planning LLM operates alongside the primary Task Planning LLM, providing safety feedback without inflating the task planner's context window. The framework introduces LLM-as-a-Judge as a novel safety metric, where a dedicated evaluator assesses plans against fifteen risk criteria including spatial conflicts, invalid action dependencies, and omitted preconditions.

Jindal et al. \cite{jindal2025can} from Google DeepMind present the most empirically rigorous evaluation of physical safety in foundation models through their ASIMOV-2.0 benchmark.  The benchmark comprises three components grounded in real-world injury narratives from the National Electronic Injury Surveillance System (NEISS) and operational safety constraints derived from ISO standards. Their evaluation of frontier models reveals troubling gaps: models show significant performance degradation when moving from text to video modalities; none achieve less than thirty percent constraint violation rates when reasoning jointly about embodiment limitations, physics, and visual cues; and smaller ``nano'' models suitable for on-device deployment perform substantially worse than their larger counterparts. The authors demonstrate that post-training with ``thinking traces'' structured reasoning about safety constraints can dramatically improve performance, with fine-tuned models achieving state-of-the-art constraint satisfaction while reducing thought length by fifty percent. This work establishes that physical safety understanding is not an emergent property of scale alone but requires targeted training on embodied scenarios.

To conclude, physical AI safety requires moving beyond the narrow, hardware-focused conceptions embedded in current standards toward multidimensional frameworks encompassing psychological, temporal, cyber-physical, and social dimensions. The integration of AI into physical systems creates novel failure modes, such as adversarial attacks on perception systems, reward hacking in reinforcement learning, out-of-distribution generalization errors etc., that traditional safety engineering cannot address alone. 
Embodiment fundamentally changes safety requirements: an AI that only processes text can cause digital harm, but an AI that controls a robot or guides a human through a repair can cause direct physical injury, demanding new benchmarks, training paradigms, and control architectures. 

\section{Robotics Physical Safety Engineering Pipeline}

\begin{figure}[ht]
\centering
\includegraphics[width = \textwidth]{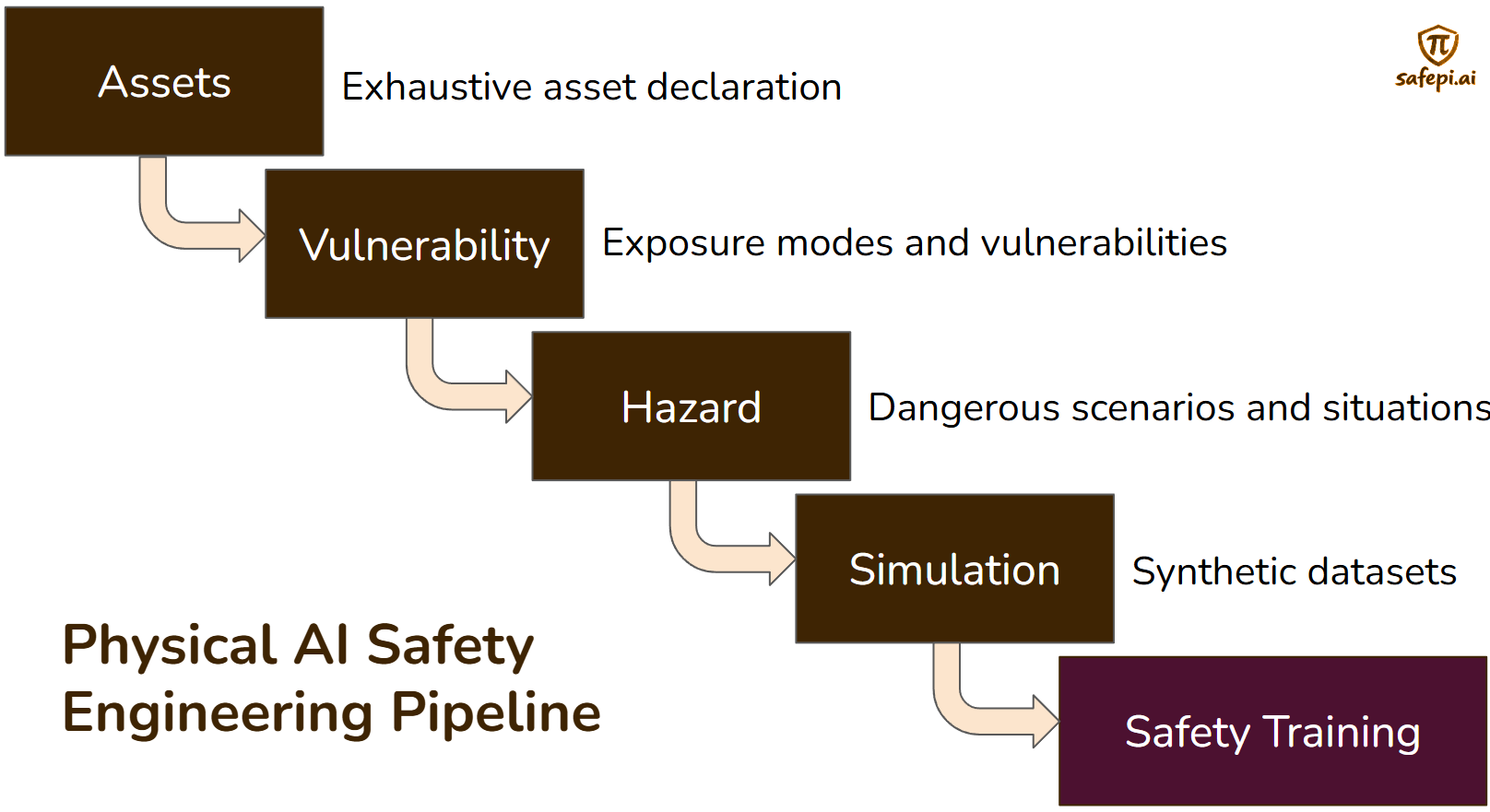}
\caption{Physical AI Safety Engineering Pipeline}
\label{fig:scheme}
\end{figure}

In this section, we propose a five-step, hazard-informed engineering pipeline. This framework systematically integrates classical risk management principles with modern machine learning workflows, leveraging synthetic data to anticipate and mitigate both localized and systemic risks. The pipeline guides practitioners from an exhaustive declaration of what must be protected, through to the fine-tuning of ML models that can actively perceive and avoid unsafe states, see Fig.~\ref{fig:scheme}.

\subsection{Step 1: Asset Declaration (Protection Universe)}

The foundation of any rigorous safety analysis is a complete and unambiguous definition of the assets to be protected. In this initial declaration phase, all potentially valuable assets must be enumerated exhaustively and at multiple levels of granularity. To avoid blind spots, no filtering or prioritization is permitted at this stage. This comprehensive inventory forms the "protection universe" for all subsequent analysis.
Examples of asset declarations include:

\begin{itemize}
\item Human assets: operator, technician, bystander

\item Sub-assets: vision, limbs, cognitive capacity, psychological safety

\item Organizational assets: robot hardware, sensors, production line, stored goods, brand reputation

\item Environmental assets: soil quality, air integrity, water systems
\end{itemize}

This step is informed by and aligns with foundational safety standards, including ISO 12100 (Safety of machinery - General principles for design - Risk assessment and risk reduction) and ISO 10218 (Robots and robotic devices - Safety requirements for industrial robots).

\subsection{Step 2: Exposure Modes (Vulnerability Enumeration)}

With the protection universe established, the next step is to define the {\em exposure modes} for each declared asset. This is an exercise in vulnerability enumeration, focusing on how an asset could become susceptible to harm. It is a taxonomy of potential weaknesses, independent of any specific cause or initiating event.
Examples of exposure modes include:

\begin{itemize}
\item Human arm: exposure to moving actuator

\item Stored goods: exposure to dropping

\item Battery pack: exposure to overheating

\item Data: exposure to corruption

\item Environment: exposure to chemical leak
\end{itemize}

This stage systematically catalogs vulnerabilities, creating a structured link between each asset and its potential points of failure.

\subsection{Step 3: Hazard Scenario Definition}

An exposure mode alone is an abstract vulnerability. The third step transforms this abstraction into concrete, testable {\em hazard scenarios} by mapping each vulnerability to specific, causal chains of events. This step introduces the context of how a vulnerability might be realized, linking system states or failures to potential harm. Examples of hazard scenarios derived from the exposure modes above include:

\begin{itemize}
\item Human in workspace: sensor occlusion causes failed detection

\item Overheating battery: cooling control loop malfunction

\item Dropping load: gripper torque miscalibration

\item Data corruption: control software overflow
\end{itemize}

This step yields a library of explicit, cause-effect scenarios that are amenable to both classical FMEA (Failure Mode and Effects Analysis) and modern simulation-based testing.

\subsection{Step 4: Simulated Scene and Synthetic Data Generation}

For every hazard scenario identified in Step 3, we generate targeted synthetic data. This is not random data augmentation but a structured generation process grounded in the hazard model. The goal is to create rich, annotated datasets that allow ML models to learn the perceptual precursors of unsafe states.

The process for each scenario involves:

\begin{itemize}
\item Digital twin construction: A high-fidelity 3D simulation environment is built to model the robot, its workspace, and relevant assets.

\item Failure mode injection: The specific failure mode from the hazard scenario (e.g., sensor occlusion, cooling malfunction) is programmatically injected into the simulation.

\item Controlled variation generation: Thousands of scene variations are automatically generated by perturbing relevant parameters, such as lighting conditions, camera viewpoints, occluding object placement, human pose, and sensor noise models.

\item Safety-relevant labeling: Each generated scene is automatically annotated with ground-truth labels critical for safety. Examples include bounding boxes for unsafe proximity, flags for imminent collision, binary indicators for overheating threshold exceeded, or semantic masks for containment breach. This creates structured datasets where the safety-relevant features are explicitly known.
\end{itemize}

This creates structured synthetic datasets grounded in explicit hazard modeling rather than random augmentation.

\subsection{Step 5: ML Fine-Tuning and Safety Envelope Learning}

The final stage integrates the hazard-informed synthetic datasets directly into the machine learning lifecycle. The objective is to imbue perception and control models with a learned understanding of the system's safety boundaries, its ``safety envelope.'' By training on data that explicitly shows the transition from nominal to hazardous states, models can be fine-tuned to not only perform their primary task but also to actively perceive and avoid risk.

This stage enables several critical capabilities:
\begin{itemize}
\item Fine-tuning perception models

\item Training anomaly detection models

\item Training hazard anticipation models

\item Stress-testing boundary cases
\end{itemize}

This five-step pipeline provides a systematic, auditable, and scalable methodology for embedding safety into the design and deployment of complex robotic systems.

\section{Practical Example: Humanoid Robot Operating in a Kindergarten}

To illustrate the framework, consider a humanoid robot deployed in a kindergarten environment. The robot assists educators by carrying objects, placing items on tables, and interacting with children.

Children move quickly, change direction unpredictably, and often lean or grab objects impulsively. A local safety policy therefore states:

\begin{quote}
\tt
Policy \#1: Any object placed on a table must be positioned\\ at least
10 cm away from the table edge.
\end{quote}

We now apply the hazard-informed pipeline step by step.

\subsection*{Step 1. Asset Declaration (Protection Universe)}

In this scenario, relevant assets include:
human assets (children, teachers and staff), organizational assets (the humanoid robot hardware, kindergarten property such as tables, floors etc.), institutional reputation and regulatory compliance.
In this context a simple object like a can becomes safety-relevant because of the human environment it is embedded in.

\subsection*{Step 2. Exposure Modes (Vulnerability Enumeration)}

For the task ``place a can on the table'', possible vulnerabilities include: exposure of a child to  falling object, exposure to liquid spill, exposure to collision if child suddenly reaches. 
A vulnerability is not yet the accident, it is the condition under which harm could become possible.

\subsection*{Step 3. Hazard Scenario Definition}

From the vulnerability ``exposure to falling object'', we derive concrete hazard scenarios:
\begin{enumerate}
 \item Robot places can 2 cm from table edge

\item Child runs past and bumps table

\item Can placed near edge falls
\end{enumerate}

The 10 cm rule acts as a risk buffer policy, a spatial safety margin to absorb environmental uncertainty and unpredictable child motion.

\subsection*{Step 4. Simulated Scene and Synthetic Data Generation}

To engineer this formally we build a 3D digital twin of the kindergarten classroom. 
The model includes table geometry and edge boundaries.
We generate scene variations to cover different table sizes, different can weights, lighting variations. 
The generated synthetic data is labeled as {\em safe placement} (more than 10 cm from edge) or {\em edge violation} (less than 10 cm).
Now the safety rule becomes computable and trainable.

\subsection*{Step 5. ML Fine-Tuning and Safety Envelope Learning}

Instead of training only for ``successful placement,'' we train models to detect table edges robustly and override task planner if placement violates the 10 cm rule.

\section{Conclusion}

This report has presented a structured, five-step hazard-informed pipeline that bridges classical risk engineering with modern machine learning workflows, providing a systematic methodology for embedding safety into the entire robotics development lifecycle.

By beginning with an exhaustive asset declaration and progressing through vulnerability enumeration to concrete hazard scenarios, the framework establishes a formal, auditable ontology of potential harm. The critical innovation lies in translating this ontology into targeted synthetic data generation, where thousands of controlled variations of hazardous situations can be simulated, annotated, and rendered into structured datasets. This transforms abstract safety rules (such as maintaining a 10cm buffer zone in a kindergarten) into computable, trainable objectives for machine learning models.

Beyond model fine-tuning, this hazard-informed synthetic data pipeline serves a dual purpose with profound implications for the certification and deployment of robots working alongside humans. The same datasets that fine-tune perception and control models can function as formal test oracles for regulatory compliance and safety validation. Regulatory bodies and certification authorities can audit not only the final model but the underlying hazard ontology and simulation fidelity that generated its training data. This creates an unprecedented layer of transparency: instead of treating a robot's safety behavior as an inscrutable black box learned from unstructured real-world data, stakeholders can examine whether the system has been explicitly trained to recognize and avoid the specific failure modes identified during risk assessment.

\bibliographystyle{vancouver}
\bibliography{ref}
\end{document}